\documentclass{article}

\usepackage{microtype}
\usepackage{graphicx}
\usepackage{subfigure}
\usepackage{booktabs} %

\usepackage{subfigure}
\PassOptionsToPackage{subfigure}{tocloft}
\usepackage{easy-todo}
\usepackage{multirow}

\usepackage{amsmath,amsfonts,bm}

\def\eqref#1{equation~\ref{#1}}

\def\1{\bm{1}}

\DeclareMathAlphabet{\mathsfit}{\encodingdefault}{\sfdefault}{m}{sl}
\SetMathAlphabet{\mathsfit}{bold}{\encodingdefault}{\sfdefault}{bx}{n}

\newcommand{\R}{\mathbb{R}}

\DeclareMathOperator*{\argmax}{arg\,max}

\usepackage{mathtools}

\newcommand{\norm}[1]{\left\lVert#1\right\rVert}
\newcommand{\fun}[3]{\ensuremath{#1\colon #2\mapsto #3}}

\newcommand{\SO}{\ensuremath{\mathbf{SO}}}

\newcommand{\tr}{\text{tr}}

\usepackage{hyperref}

\usepackage[accepted]{icml2019}

\icmltitlerunning{Cross-Domain 3D Equivariant Image Embeddings}

\begin{document}

\twocolumn[
\icmltitle{Cross-Domain 3D Equivariant Image Embeddings}

\icmlsetsymbol{equal}{*}

\begin{icmlauthorlist}
\icmlauthor{Carlos Esteves}{equal,penn}
\icmlauthor{Avneesh Sud}{goo}
\icmlauthor{Zhengyi Luo}{penn}
\icmlauthor{Kostas Daniilidis}{penn}
\icmlauthor{Ameesh Makadia}{goo}
\end{icmlauthorlist}

\icmlaffiliation{penn}{GRASP Laboratory, University of Pennsylvania}
\icmlaffiliation{goo}{Google Research}

\icmlcorrespondingauthor{Carlos Esteves}{machc@seas.upenn.edu}

\icmlkeywords{equivariance, cross-domain, spherical cnn, pose estimation, novel view synthesis}

\vskip 0.3in
]

\printAffiliationsAndNotice{\textsuperscript{*}Work done during an internship at Google.}  %

\begin{abstract}
Spherical convolutional networks have been introduced recently as tools to learn powerful feature representations of 3D shapes. Spherical CNNs are equivariant to 3D rotations making them ideally suited to applications where 3D data may be observed in arbitrary orientations. In this paper we learn 2D image embeddings with a similar equivariant structure: embedding the image of a 3D object should commute with rotations of the object.  We introduce a cross-domain embedding from 2D images into a spherical CNN latent space. This embedding encodes images with 3D shape properties and is equivariant to 3D rotations of the observed object. The model is supervised only by target embeddings obtained from a spherical CNN pretrained for 3D shape classification. We show that learning a rich embedding for images with appropriate geometric structure is sufficient for tackling varied applications, such as relative pose estimation and novel view synthesis, without requiring additional task-specific supervision.%
\end{abstract}

\section{Introduction}

\begin{figure*}
\centering
 \includegraphics[width=\linewidth]{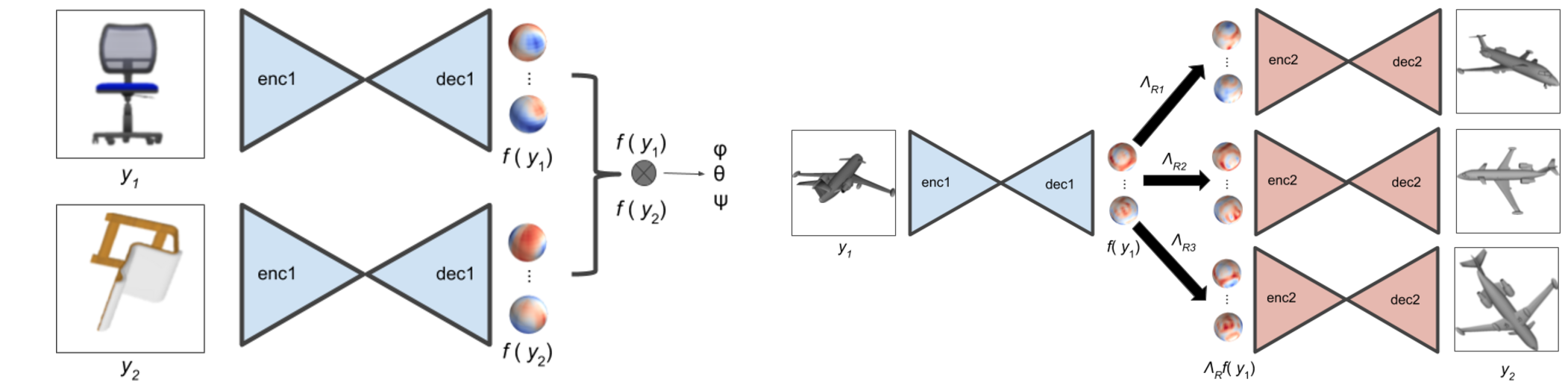}%
\caption{
  \textbf{Overview.} We learn category based spherical 3D equivariant embeddings that can be correlated for relative pose estimation, and rotated for novel view synthesis.
  \textit{Left: relative pose estimation}.
  Given 2 images of objects from same class, we obtain the respective spherical embeddings.
  The relative pose is computed from the spherical correlation between the spherical embeddings.
  \textit{Right: novel view synthesis.}
  We first embed the input view into the spherical representation,
  then we apply the target rotation to the spherical feature maps, and feed them to the synthesizer to generate novel views.
}
\label{fig:pose_synth}
\end{figure*}
The success of CNNs in computer vision has shown that large training datasets and task-specific supervision are sufficient to learn rich feature representations for a variety of tasks such as image classification and object detection~\cite{he16_resnetv2}. However, there remain many challenges, such as motion estimation and view synthesis, which require complex geometric reasoning and for which labeled data is not available at scale. For such problems there is a trend towards developing models with geometry-aware latent representations that can learn the structure of the world without requiring full geometric supervision (e.g.~\citet{kulkarni15nips,rhodin18eccv,yang2015weaklysupervised,yan16nips,mahjourian18cvpr,Eslami1204}).

A desirable property for an image embedding is robustness to 3D geometric transformations of the scene. 3D rotations, in particular, are a nuisance to computer vision algorithms because even small 3D rotations of objects in the world can induce large transformations in image space.
In recent years there has been much attention given to the study of equivariant neural networks (e.g. \citet{cohen2016group,worrall2017harmonic}), as equivariant maps provide a natural formulation to address group transformations on images.  Despite these advances, designing a 3D rotation equivariant map of 2D images is an open challenge. This is because the rotation of a 3D object does not act directly on the pixels of the resulting image due to the intervening camera projection. Thus, an equivariant map cannot be constructed by design and instead an (approximate) equivariant map must be learned. This is the central task of the paper: \emph{how can we learn an embedding for images of 3D objects that is equivariant to 3D rotations of the objects?}

Our proposal is inspired by recent works on 3D rotation equivariant CNNs for 3D shape representations~\cite{s.2018spherical,esteves18eccv}. These works show that spherical convolutional networks can achieve state of the art performance on 3D shape classification and pose estimation tasks, and these networks' equivariance properties mean their performance does not suffer when considering 3D shapes in arbitrary orientations. 

In this work we propose to learn an equivariant embedding of an image by mapping it into the equivariant feature space of a spherical CNN trained on datasets of 3D shapes. Our approach is unique in that we are directly supervising the desired target embeddings with the pretrained 3D shape features and we do not consider any other task-specific training losses. By bootstrapping with features of 3D shapes, our model (1) encodes images with the shape properties of the observed object and (2) has an an underlying spherical structure that is equivariant to 3D rotations of the object.

The cross-domain embeddings can be used for different applications, either directly or indirectly, without requiring any additional task-specific supervised training. We illustrate this point by showing results on two very different challenges:
\paragraph{Relative orientation estimation}
Our model maps images to rotation equivariant embeddings defined on the sphere  (Fig.~\ref{fig:pose_synth}-\textit{left}). We can recover the relative orientation between two embeddings by finding the rotation that brings them into alignment. We do this simply with spherical cross-correlation without running any regression as an unsupervised spatial transformer~\cite{jaderberg15nips} would do. This method approaches state of the art even though it uses no task-specific training. The same method can be applied to align 2D images with 3D shapes.
\paragraph{Novel view synthesis}
The learned embeddings also encode enough shape properties to synthesize new views. By simply training a decoder from the spherical embedding space with a photometric loss, we have a model for novel view synthesis. New views are generated by rotating the latent embedding (Fig.~\ref{fig:pose_synth}-\textit{right}). No task specific supervision (e.g. an image and its rotated counterpart) is required.

To reiterate, our main contribution is a novel cross-domain neural model that can map 2D images into a 3D rotation equivariant feature space. Generating spherical feature maps from 2D images is a complex high-dimensional regression task, mapping between topologies, which requires a novel encoder-decoder architecture. We consider the relative pose and view synthesis tasks as proxies for analyzing the representation power of our learned embeddings. Nonetheless, our promising experimental results indicate our cross-domain embeddings may be useful for a variety of tasks.

\section{Related Work}
\label{sec:related_work}
A number of recent works have introduced geometric structure to the feature representations of deep neural networks. The most common setting is to learn intermediate features that can be directly manipulated or transformed for a particular task. For example, in ~\citet{rhodin18eccv}, \citet{worrall2017iccv}, \citet{cohen15iclr}, \citet{hinton11icann}, \citet{yang2015weaklysupervised}, and \citet{kulkarni15nips},  geometric transformations can be directly applied to image features (in some cases disentangled pose features), in order to synthesize new views. In a related approach, ~\citet{tatarchenko16eccv} uses an encoder-decoder architecture that augments pose information to the latent image embedding. One drawback of these methods is that they typically require full supervision, where both the geometric transformation parameters and the corresponding target image are available as supervision during training. 
Furthermore, training with source-target pairs requires covering a large sample space - synthesizing views from arbitrary relative 3D orientations requires sampling pairs from $\SO(3) \times \SO(3)$. In contrast, our model trains with a single image per example.

Different to all of the methods above, Homeomorphic VAEs \cite{falorsi18icmlw} provide an unsupervised way to learn an $\SO(3)$-latent-embedding for images. However, it is presently unclear if this method can scale to practical scenarios (it requires a dense sampling of views to learn a continuous embedding, dealing with intra-class variations, etc).

While the examples above have all been applied towards the task of view synthesis, there also exist a variety of other approaches to this problem. Most relevant to our setting are the self-supervised methods that learn geometrically meaningful embeddings using differentiable rendering to match semantic maps~\cite{yao20183d}, shading information~\cite{henderson2018learning}, fusing latent embeddings from multiple views, and improving synthesis using multiple rendering steps~\cite{Eslami1204}. 

\textbf{Pose Estimation:} The task of object pose estimation has been a long standing problem with numerous applications in computer vision and robotics. Most approaches can be categorized as keypoint-based or direct pose estimation as regression or classification. Keypoint-based methods for object pose estimation include \citet{pavlakos17object3d} and \citet{grabner18arxiv}, the former predicting semantic keypoints and the latter bounding box corners, from which object pose is determined from a PnP algorithm.
Direct pose estimation methods include \citet{vpsKpsTulsianiM15} and \citet{su15iccv} where classification is performed over a quantized viewpoint space. \citet{kanezaki2018cvpr} train a joint 3D object classification and pose CNN from multiple views with unknown viewpoints, however the viewpoint sphere is sampled discretely providing limited resolution in the estimated pose.
\citet{mahendran17cvprw} introduces a carefully designed CNN for viewpoint regression, analyzing different representations and geodesic loss functions, and \citet{MousavianCVPR17} introduce a MultiBin orientation regression network. KeypointNet~\cite{suwajanakorn2018discovery} learns category-specific semantic keypoints and their detectors using only a geometric loss. The 3D keypoints are also useful for determining relative pose, although the method struggles when exposed to arbitrary 3D rotations due to lack of rotation equivariance.

The key ingredient in our approach is a novel method to map 2D images to rotation-equivariant 3D shape embeddings, essentially encoding an image with 3D geometric structure. We note that the choice of geometric representation (spherical embeddings) is intentional in order to maintain rotation equivariance. Alternative geometric representations such as volumetric (e.g. single-view volumetric reconstruction from \citet{tulsiani17cvpr}) would not be rotation equivariant (although \citet{weiler3dsteerable} could provide an alternative for certain tasks).

\section{Method}
In this section we detail our image embedding model. We begin by revisiting spherical CNNs (Section~\ref{sec:sphericalcnn}) as a means to learn rich equivariant embeddings for 3D shapes, and Section~\ref{sec:basic} introduces our cross-domain architecture that learns to map 2D images into the same embedding space. Sections~\ref{sec:relpose_subsec} and ~\ref{sec:nvs_subsec} will describe how these image embeddings can be used for relative pose estimation and novel view synthesis.
\subsection{Spherical CNNs}
\label{sec:sphericalcnn}
Spherical CNNs~\cite{s.2018spherical,esteves18eccv} produce 3D rotation equivariant feature maps for inputs defined on the sphere. In practice these methods have been useful for a variety of 3D shape analysis tasks as it is common for inputs to appear in arbitrary pose, for which equivariance is a particularly helpful property. In this work we adopt the spherical convolutional model introduced in~\citet{esteves18eccv} due to its efficiency and performance on 3D shape alignment tasks\footnote{It is important to note that we are tackling the more challenging problem of relative 3D pose from 2D images.}. We briefly summarize the spherical CNN below (see~\citet{esteves18eccv} for details). For functions $x_1, x_2$ defined on the sphere, their convolution is defined as
\begin{align}
(x_1 \star x_2)(p) = \int_{R \in \mathbf{SO}(3)} x_1(R\eta)x_2(R^{-1}p)dR,
\end{align}
where $\eta$ is the north pole of the sphere (the stationary point under \SO(2)). This extends to K-channel inputs in a straightforward manner
\begin{align}
(x_1 \star x_2)(p) = \sum_{k=0}^{K-1} \int_{R \in \mathbf{SO}(3)} x_{1,k}(R\eta)x_{2,k}(R^{-1}p)dR,
\end{align}
where $x_{\cdot,k}$ denotes the $k$-th channel.

These convolutions are the primary building blocks of spherical CNNs. 
We define $s$ as a spherical CNN that maps $K_{\text{in}}$-channel spherical inputs to $K_{\text{out}}$-channel spherical feature maps.
Precisely, in the single-channel case we have $\fun{s}{L^2(S^2)}{L^2(S^2)}$ where $L^2(S^2)$ denotes square-integrability, which is necessary for the efficient evaluation of convolution in the spectral domain.

An important property of the spherical CNN is 3D rotation equivariance. For any \fun{x}{S^2}{\R^{K_\text{in}}}, we have
\begin{align}
 s(\Lambda_{R} x) =  \Lambda_{R} s(x),
 \label{eq:s_equiv}
\end{align}
where $\Lambda_R$ is the rotation operator by $R\in \SO(3)$%
\footnote{We use $\Lambda_R$ as a generic rotation operator that can be applied to 3D shapes and spherical functions, scalar or vector-valued. Interpretation should be clear from context.}.
Technically this equivariance is only approximate as the nonlinear activations (ReLU) and spatial pooling operations break the bandlimiting assumptions which otherwise guarantee equivariance. However, in practice these errors are negligible~\cite{esteves18eccv}.

To use spherical CNNs with 3D shapes, we must provide a map $r(M)$ which converts any 3D shape $M$ to a spherical representation. While there are many choices for $r$ we use the simple ray-casting technique of~\citet{esteves18eccv}. Most importantly, $r(M)$ is equivariant to 3D rotations which ensures end-to-end equivariance of our 3D shape feature maps: $s(r(\Lambda_{R} M)) = \Lambda_{R}s(r(M))$.

\subsection{Cross-domain spherical embeddings}
\label{sec:basic}
In the previous section we summarized a rotation equivariant spherical CNN model for 3D shape inputs: $s(r(M))$. The primary objective of this work is to learn an \emph{image} embedding that can capture similar underlying 3D shape properties and equivariant structure.  Specifically, let us define an RGB image as the projection $c$ of a shape $M$ ($c$ can be any usual camera projection model, e.g. perspective or orthographic). We seek a map $f(c(M))$ that captures the shape properties of $M$ and retains an equivariant structure: $f(c(\Lambda_R M)) = \Lambda_R f(c(M))$. This is challenging because $c(M)$ is not 3D rotation equivariant as it is a camera projection, so we cannot have equivariance by construction. We propose to learn an approximately equivariant embedding model $f$ using a spherical CNN for 3D shapes, i.e. a pretrained $s(r(M))$), as supervision: we wish to learn $f$ such that $f(c(M)) = s(r(M))$. If learned successfully, the equivariance of $f$ follows simply from (\ref{eq:s_equiv}):
\begin{align}
	f(c(\Lambda_R M)) &= s(r(\Lambda_R M)) \nonumber \\
                    &= \Lambda_R s(r(M))  \\
                    &=\Lambda_R f(c(M)). \nonumber
    \label{eq:img_equiv}
\end{align}
Since $c(M)$ and $r(M)$ are fixed and not part of the trainable model, we substitute $y=c(M)$ and $x=r(M)$ going forward to simplify notation.

Learning $f$ involves predicting high dimensional multi-channel spherical maps from a single image.
The two major design challenges are deciding the structure of $f(y)$ and the training loss $\mathcal{L}(x, y)$ from predicted embedding $f(y)$ to the target ground truth $s(x)$. We describe first the training loss.
For simplicity we describe the loss for a single channel (in general the loss is  aggregated over the channels).
The implementation of the spherical CNN represents the spherical function $s(x)$ on a grid via equirectangular projection. A discretized $s(x)$ of resolution $N \times N$ can be indexed by the pair $(\theta_i, \phi_j), i,j \in \left\{0,1,...,N-1\right\}$. The $\theta_i$ uniformly sample colatitude, and similarly $\phi_j$ uniformly sample azimuth.
Since our target embeddings are unbounded, we found crucial to use a robust loss such as Huber%
\footnote{median pose errors are $\approx 10$ deg larger with $L_1$ or $L_2$}, and a Huber breakpoint at 1 works well in practice.
The loss follows, where $\mathcal{H}$ is the Huber loss, and a weight is introduced to account for the nonuniform equirectangular spherical sampling ($\sin(\theta)$ is proportional to the sample area):
 \begin{align}
 \mathcal{L}(x, y) &= \frac{1}{N^2} \sum_{i,j = 0}^{N-1} \mathcal{H}(\sin(\theta_i)(f(y)-s(x))(\theta_i, \phi_j)) \\
  \mathcal{H}(\alpha) &= \begin{cases}
    0.5 \alpha^2                   & \text{for } |\alpha| \le 1, \\
    |\alpha| - 0.5 & \text{otherwise.}
  \end{cases}  \label{eq:loss}
\end{align}

\subsubsection{Architecture}
We now describe the structure of our cross-domain embedding model $f(y)$.
With $f(y)$ we are predicting spatially dense spherical feature maps from a single 2D image. 
Convolutional encoder-decoder architectures with skip connections such as U-Net~\cite{ronneberger15miccai} or Stacked Hourglass~\cite{newell2016stacked} produce excellent results when some pixelwise association can be made between the input and output domains (e.g. for dense labeling tasks like semantic segmentation~\cite{deeplabv3plus2018}). However, we must learn a cross-domain map from 2D image (Euclidean) to spherical functions.  In this setting, architecture features such as skip connections are not only unnecessary but can be harmful by forcing the network to incorrectly consider associations across topologies.\footnote{Although cross-modal learning has been explored in different domains, e.g. \citet{aytar2016soundnet}, these methods predict representations in $\mathbb{R}^n$ from different modalities which is a simpler application of 1D and 2D CNNs.}

We consider an encoder-decoder architecture, with several rounds of downsampling from input image to a 1D vector, followed by rounds of upsampling from the 1D vector to the set of spherical feature maps.
We follow the best practices for this kind of architecture proposed by \citet{dcgan}, employing a fully convolutional network with strided convolutions for downsampling and transposed convolutions for upsampling.
We apply azimuthal circular padding after the 1D bottleneck, when the feature maps are expected to assume spherical topology.
We also found performance improvements by replacing convolutional layers with residual layers \cite{he16_resnetv2}. 
Figure~\ref{fig:architecture} shows the architecture.
See supplementary material for more details.

\subsubsection{Target embeddings}
The remaining decision is to select the appropriate target feature maps from $s(x)$. For all our experiments $s(x)$ is a 10 layer residual spherical CNN trained for ModelNet40 3D shape classification on $64 \times 64$ inputs (i.e. $r(M)$ produces a single-channel $64 \times 64$ output). The decision of which feature maps to use as the target is application-dependent. For category-based relative pose estimation, we want features that are void of instance level details, which is achieved by taking the target embedding from deeper layers. For view synthesis, we wish that the instance-level details are preserved, so we embed to an earlier layer. We employ the same pre-trained spherical CNN for all  experiments (on ModelNet40, ObjectNet3D and ShapeNet), which shows generalization; more details in the supplementary material.

\begin{figure}[ht]
\centering
\includegraphics[width=\linewidth]{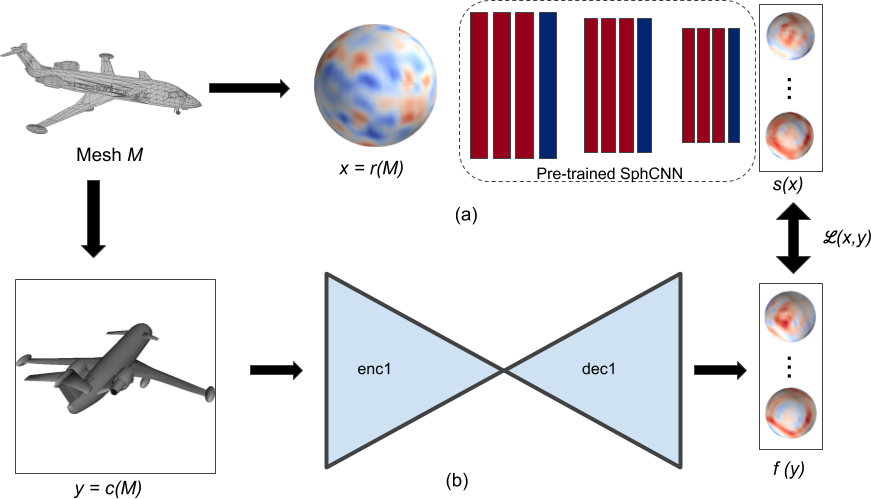}%
\caption{
  \textbf{Cross-domain spherical embeddings.} Given a 3D mesh, (a) we map it to a spherical function, and use a pre-trained spherical CNN to compute its spherical embedding. (b) During training, we render a view and learn the transformation to the target spherical embedding using an encoder-decoder. For inference, the inputs are 2D images and only the encoder-decoder part is used.
}
\label{fig:architecture}
\end{figure}

\subsubsection{Relative pose estimation}
\label{sec:relpose_subsec}
The cross-domain embeddings produced by $f(y)$ are sufficient to recover the relative pose between pairs of images (even between different instances of the same object category). As $f(y)$ produces spherical feature maps that have been trained to be 3D rotation equivariant, we can apply 3D rotations directly to the feature maps. Relative orientation estimation is simply identifying the rotation that brings feature maps into alignment. For alignment we use a very simple cross-correlation measure. Given two images $y_1, y_2$ we estimate their relative pose as
\begin{align}
\argmax_{R\in \SO(3)} G(R) &= \sum_{k=0}^{K-1} \int_{p\in S^2} f(y_1)_k(p) \cdot f(y_2)_k(R^Tp) dp 
\label{eq:corrargmax}
\end{align}
Here the subscript $k$ denotes the $k$-th spherical channel in the image embedding. $G(R)$ can be evaluated efficiently in the spectral domain (similar in spirit to spherical convolution, see~\citet{kostelec2008ffts,makadia2010spherical} for details and implementation).

The resolution of $G(R)$ depends on the resolution of the input spherical functions $f(y_1)$ and $f(y_2)$. Our learned feature maps have a spatial resolution of $16 \times 16$ which corresponds to a cell width of 22.5 deg at the equator, which we consider too coarse for precise relative pose. To increase resolution, we upsample the features by a factor of 4 using bicubic interpolation prior to evaluating~(\ref{eq:corrargmax}).

This method can also be applied to estimate relative pose between an image $y$ and mesh $M$, by computing the argmax correlation~(\ref{eq:corrargmax}) between $f(y)$ and $s(r(M))$.

Recall, during training we take as input arbitrarily oriented meshes. A  training example consists of only the target embeddings from the pretrained $s(r(M))$ and a single view rendered from a fixed camera $c(M)$. No orientation supervision is required, and the model never sees pairs of images together at training. This reduces the sample complexity and leads to faster convergence.

\begin{figure}[h]
\centering
\includegraphics[width=\linewidth]{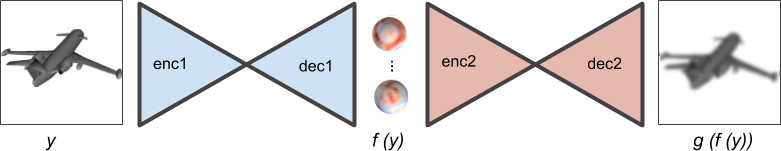}%
\caption{
  \textbf{Novel view synthesis training.}
  We learn the inverse map from spherical embeddings to 2D views.
  The map from 2D view to spherical embeddings (in blue) is the same as in Fig.~\ref{fig:architecture} and is frozen during training.
  The synthesizer network (in red) reconstructs the same input view and is trained with an $L_2$ loss.
}
\label{fig:synthtrain}
\end{figure}

\begin{table*}[ht]
  \centering
  \newcommand{\B}[1]{\textbf{#1}}
  \footnotesize
  \begin{tabular}{|l|l|l|rrr|rrr|rrr|rrr|}
    \hline
    \multicolumn{3}{|l|}{} & \multicolumn{3}{c|}{airplane} & \multicolumn{3}{c|}{car} & \multicolumn{3}{c|}{chair} & \multicolumn{3}{c|}{sofa}                                                  \\

    \multicolumn{3}{|l|}{} & med.                & a@15  & a@30     & med.     & a@15     & a@30     & med.     & a@15     & a@30     & med.     & a@15     & a@30                           \\
    \hline
    \multirow{6}{*}{2DOF}  & \multirow{3}{*}{IB} & Ours  & \B{5.17} & \B{85.3} & \B{91.9} & \B{3.70} & \B{92.2} & 92.5     & \B{5.07} & \B{90.6} & \B{94.1} & \B{4.59} & \B{93.6} & \B{95.2} \\
                           &                     & Regr. & 16.9     & 46.3     & 68.7     & 6.55     & 83.5     & \B{93.1} & 13.7     & 53.9     & 78.3     & 17.3     & 43.2     & 69.4     \\
                           &                     & KpNet & 6.95     & 79.4     & 91.5     & div.     & div.     & div.     & 6.34     & 84.7     & 91.8     & 9.20     & 71.3     & 85.4     \\
    \cline{2-15}
                           & \multirow{3}{*}{CB} & Ours  & \B{6.24} & 79.0     & 88.2     & \B{4.73} & 73.2     & 73.3     & 12.1     & 59.3     & 74.4     & \B{10.8} & \B{58.7} & 70.5     \\
                           &                     & Regr. & 20.6     & 38.7     & 63.7     & 7.06     & \B{82.4} & \B{92.5} & 16.8     & 43.7     & 72.0     & 19.6     & 37.8     & 66.5     \\
                           &                     & KpNet & 9.07     & \B{79.4} & \B{91.5} & div.     & div.     & div.     & \B{8.07} & \B{79.5} & \B{90.2} & 15.1     & 49.8     & \B{71.8} \\
    \hline
    \multirow{6}{*}{3DOF}  & \multirow{3}{*}{IB} & Ours  & \B{6.64} & \B{80.9} & \B{91.9} & \B{3.84} & \B{97.3} & \B{98.8} & \B{5.55} & \B{89.1} & \B{95.7} & \B{5.21} & \B{90.4} & \B{94.8} \\
                           &                     & Regr. & 45.4     & 12.6     & 31.3     & 9.83     & 69.0     & 86.5     & 21.7     & 31.3     & 64.3     & 22.2     & 34.8     & 61.4     \\
                           &                     & KpNet & 14.9     & 50.3     & 76.6     & 9.12     & 70.4     & 80.9     & 10.8     & 66.7     & 85.3     & 25.0     & 27.4     & 57.3     \\
    \cline{2-15}
                           & \multirow{3}{*}{CB} & Ours  & \B{7.27} & \B{76.4} & \B{89.4} & \B{4.59} & \B{92.1} & \B{93.3} & \B{12.3} & \B{59.5} & 77.3     & \B{9.66} & \B{63.9} & \B{76.0} \\
                           &                     & Regr. & 44.4     & 14.1     & 32.1     & 10.5     & 66.5     & 85.6     & 25.6     & 25.1     & 57.2     & 24.5     & 30.9     & 58.1     \\
                           &                     & KpNet & 16.3     & 46.0     & 75.0     & 10.7     & 64.4     & 77.6     & 13.6     & 55.4     & \B{81.6} & 37.4     & 12.7     & 39.8     \\
    \hline
  \end{tabular}%
  \caption{\textbf{ShapeNet relative pose estimation results.}
    We show median angular error in degrees  (\emph{med.}), accuracy \emph(a@) at 15 and 30 deg for instance (\emph{IB}) and category-based (\emph{CB}), 2 and 3 degrees of freedom relative pose estimation from single views on ShapeNet.
    Comparison is against \citet{mahendran17cvprw} (\emph{Regr.}) and \citet{suwajanakorn2018discovery} (\emph{KpNet}).
    KeypointNet does not converge on the full 3DOF setting; we limit the viewpoints to a hemisphere when evaluating it.
  Note that we still outperform it.}
  \label{tab:shapenet}
\end{table*}

\subsubsection{Novel view synthesis}
\label{sec:nvs_subsec}
The spherical embeddings learned by our method can also be applied towards novel view synthesis. 
The rotation equivariant spherical CNN feature maps undergo the same rotation as its inputs, so if we learn the inverse map that generates an image back from its embedding, we can rotate the embeddings and generate novel views. 

We define the inverse map $g=f^{-1}$ such that $g(f(y)) = y$. If we let $y_1 = c(M)$ and $y_2 = c(\Lambda_R M)$ (i.e. $y_{1,2}$ are images of shape M before and after it undergoes a 3D rotation, respectively).
It follows that
\begin{equation}
  g(\Lambda_R f(y_1)) = g(f(y_2)) = y_2.
\end{equation}

This gives us a way to generate a novel view of the 3D object under rotation, from the spherical embedding of a single view as follows (see Fig.~\ref{fig:pose_synth} for illustration):
\begin{enumerate}
    \item Obtain the embedding $f(y_1)$ of given view $y_1$,
    \item Rotate the embedding by the desired $R \in \SO(3)$, obtaining $f(y_2) = \Lambda_R f(y_1)$,
    \item Apply $g$ to obtain the novel view $y_2=g(f(y_2))$.
\end{enumerate}

Since $g$ is learning the inverse of $f$, we similarly design $g$ as a convolutional encoder-decoder, which is trained from single views enforcing $g(f(y)) = y$ with a pixel-wise $L_2$ loss $\mathcal{L}_s(y) = \norm{g(f(y))  - y}_2^2$  (see Fig.~\ref{fig:synthtrain} for illustration).

\section{Experiments}

\begin{figure*}[t]
\centering
\includegraphics[width=\linewidth]{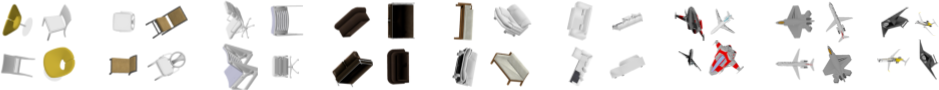}%
\caption{
\textbf{Category-based relative pose estimation.}
We render one object in the pose of the other using our estimated relative pose.
\textit{For each block, top:} Inputs 1 and 2, from the test set.
\textit{Bottom:} Mesh 2 rotated into pose 1, and mesh 1 rotated into pose 2.
We render from the ground truth meshes for visualization purposes only; the inputs to our method are solely the 2D views and the output is the relative pose.
Note how the alignment is possible even under large appearance variation.
}
\label{fig:align}
\end{figure*}

\begin{figure*}[t]
\centering
\includegraphics[width=\linewidth]{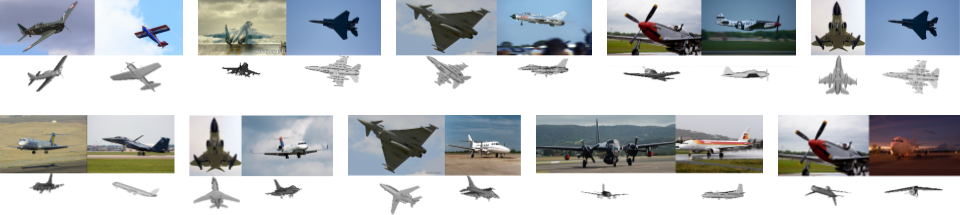}%
\caption{
\textbf{Relative pose estimation for real images.}
We render the mesh corresponding to one input in the pose of the other using our estimated relative pose.
\textit{For each block, top:} Inputs 1 and 2, from the test set.
\textit{Bottom:} Mesh 2 rotated into pose 1, and mesh 1 rotated into pose 2.
Image pairs on the top row map to the same mesh in the dataset; on the bottom they map to different meshes.
The bottom-right block shows a typical failure case due to symmetry. 
Meshes are used for visualization purposes only; the inputs to our method are the 2D images and the relative pose is estimated directly from their embeddings via cross-correlation (see text for details).
}
\label{fig:o3d-align}
\end{figure*}

\label{sec:experiments}
\subsection{Datasets}
We utilize the standardized large datasets of 3D shapes ModelNet40 \cite{Wu20153Dshapenets} and ShapeNet \cite{shapenet2015} for most of our experiments.

Some methods must explicitly deal with the symmetries present in many shape categories (e.g. \citet{saxena09icra,rad17iccv}).
Our method is immune to this problem by not requiring pose annotations.
However, pose annotations are used for evaluation, therefore we limit our experiments to categories which are largely free of symmetry and thus for which relative orientation is unique.

Symmetry is a problem for the evaluation of ShapeNet \emph{airplanes}.
Some of the instances (e.g. spaceships and flying wings) are fully symmetric around one axis, which results in non-injective embeddings and two possible correct alignments that differ by 180 deg.
For meaningful evaluation we compute the errors up to symmetry for this class.

Recall that we are not estimating pose relative to a canonical object frame but rather relative object orientation from a pair of images. Thus, for training, we do not require that our dataset models come aligned per category, and in fact we introduce random rotations at training time. For evaluation, in order to quantify our inter-instance performance we require aligned shapes to determine the ground truth (see Section~\ref{sec:rel-pose-experiments}); for ModelNet40 we use the aligned version from \citet{sedaghat15iccv}.

Multiple datasets have been proposed for object pose estimation, such as Pascal3D+ \cite{xiang14wacv}, KITTI \cite{geiger12cvpr}, and Pix3D \cite{pix3d}, but they do not exhibit large variation in viewpoints, especially camera elevation. For example Pascal3D+ has most elevations concentrated inside $\left[-10^\circ, 10^\circ\right]$ and the official evaluation only considers azimuthal accuracy. In our setting we explore geometric embeddings that can capture more challenging arbitrary viewpoints. Our results indicate that the problem of relative orientation from two views is quite difficult even for rendered images from ModelNet40 and ShapeNet.
Our experiments with real images are limited to the \emph{airplane} and \emph{cars} categories of ObjectNet3D \cite{xiang2016objectnet3d}, which have the largest variety of viewpoints among all categories.

\subsection{Relative pose estimation}
\label{sec:rel-pose-experiments}
For training, we render views in arbitrary poses sampled from \SO(3).
We have two modes of evaluation: \textit{instance} and \textit{category} based. 
For category-based, we measure the relative pose error between each instance and 3 randomly sampled instances from the test set. 
For instance-based, we measure the error between each instance from the test set and 3 randomly rotated versions of itself. 
The error is the angle between the estimated and ground truth relative poses;
given input ground truth poses $R_1$ and $R_2$ and estimated pose $R$, it is given by $\arccos\left({(\tr{(R_2^\top R_1 R)}-1)/2}\right).$
We compare with the following methods.

\textbf{Regression:}
We consider a method based on \citet{mahendran17cvprw}, which treats pose estimation as regression.
To keep the comparison fair we use approximately the same number of parameters in the encoder as in our networks.
See supplementary material for more details.
\citet{mahendran17cvprw} requires the ground truth pose with respect to a canonical orientation during training, whereas our method is self-supervised and can operate on unaligned meshes.
We still outperform it even when allowing extra information, especially in the presence of 3DOF rotations.

\textbf{KeypointNet:}
\citet{suwajanakorn2018discovery} introduce an unsupervised method of learning keypoints that can be used for pose estimation.
Similarly to our method, it generates training data by rendering different views from meshes.
It requires consistently oriented meshes for dominant direction supervision, whereas our method makes no assumptions about mesh orientation.
While they show results for 2DOF rotations, only viewpoints on a hemisphere are considered, whereas we sample the whole sphere.
We retrain and evaluate KeypointNet with full 2DOF and 3DOF rotations.
Our method outperforms it on the more challenging scenarios.

Table~\ref{tab:shapenet} shows ShapeNet results.
Figure~\ref{fig:align} shows the 3DOF alignment quality on ShapeNet by rendering views using the estimated relative poses.
We show results for ModelNet40 and for aligning meshes to images in the supplement.

\subsubsection{Extension to real images}
Most labeled real-world object pose estimation datasets have restricted pose variations. One dataset with sufficient variation of 3D poses is the airplane class in ObjectNet3D~\cite{xiang2016objectnet3d}.
We assume object instance bounding boxes are provided (e.g. using an object detection network~\cite{cvpr17_object}).
We also experiment with the \emph{cars} category by augmenting it with in-plane rotations to increase the pose variation.
We train our model on image-mesh pairs and significantly outperform the method based on regression. \emph{airplanes} numbers are up to a 180 deg rotation due to symmetry as explained in Section \ref{sec:discuss} (see bottom right of Figure~\ref{fig:o3d-align} for an example).
Table~\ref{tab:shapenet} shows the comparison while Figure~\ref{fig:o3d-align} shows alignment results for \emph{airplanes}.
\begin{table}[ht]
\centering
\begin{tabular}{|l|l|rrr|}
\hline
\multicolumn{2}{|c|}{}      & med err. & acc@15         & acc@30                          \\
  \hline
  \multirow{2}{*}{airplane} & Ours        & \textbf{13.75} & \textbf{53.40} & \textbf{76.60} \\
                            & Regression  & 36.52          & 16.70          & 40.40          \\
\hline
\multirow{2}{*}{car}        & Ours        & \textbf{8.22}  & \textbf{72.51} & \textbf{78.00} \\
                            & Regression  & 16.16          & 46.87          & 74.35          \\
\hline
\end{tabular}%
\caption{Relative pose estimation results for real images from ObjectNet3D.
  We show median angle error in degrees and accuracy at 15 and 30 deg.
We significantly outperform the regression method based on \citet{mahendran17cvprw}.}
\label{tab:shapenet}
\end{table}

\subsection{Novel view synthesis}
\begin{figure*}
\centering
\includegraphics[width=0.9\linewidth]{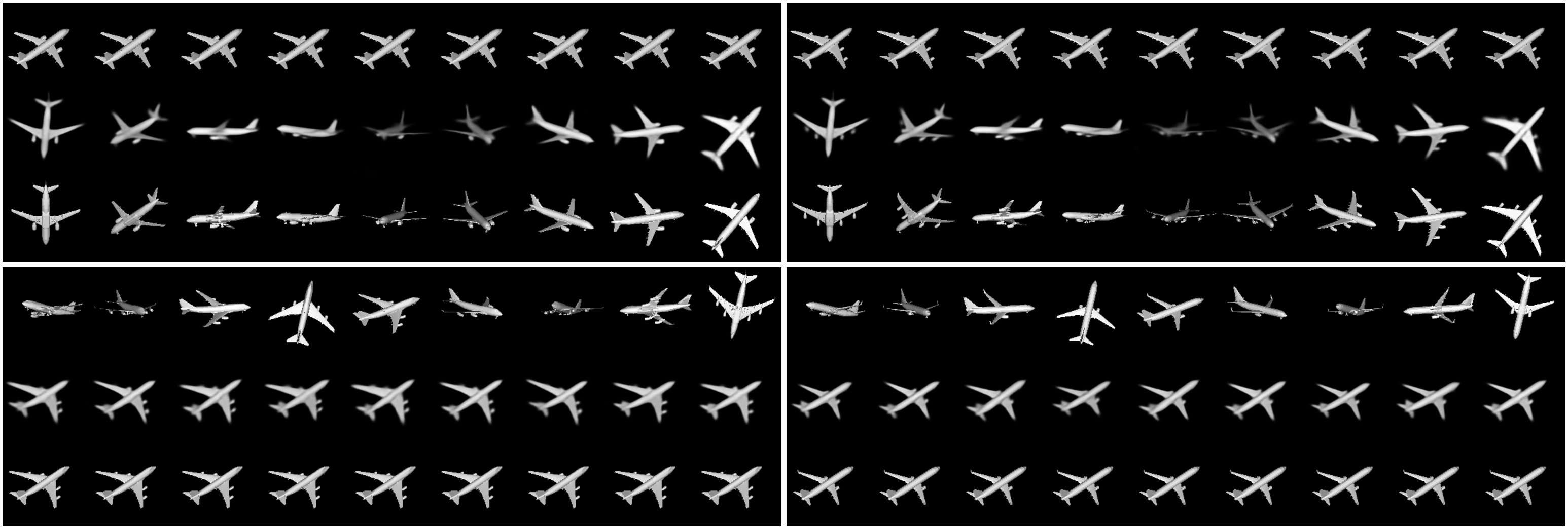}%
\caption{
  \textbf{Novel view synthesis.}
  Our embeddings are category based, capture both geometry and appearance, can be rotated as spheres, and can be inverted through another neural network.
  We can generate any new viewpoint from any given viewpoint.
  For each block: \textit{top row:} inputs; \textit{middle row:} novel views generated using our method; \textit{bottom row:} ground truth views rendered from the original mesh.
  Top two blocks show different views generated from a single image; bottom two blocks show a single view generated from different images.
}
\label{fig:novel}
\end{figure*}

We evaluate novel view synthesis qualitatively%
\footnote{We attempted a method similar to \citet{tatarchenko16eccv} as baseline, with and without adversarial losses, but results were poor for our large space of rotations.}.
Figure~\ref{fig:novel} shows the results for several generated views in different poses, with a single 2D image as input. 
We do not expect to generate realistic images here, since the embeddings do not capture color or texture and the generator is trained with a simple $L_2$ loss.
Our goal is to show that the learned embeddings  naturally capture the geometry, which is demonstrated by this example, where a simple 3D rotation of the spherical embeddings obtained from a single 2D image produces a novel view of the corresponding 3D object rotation. Adversarial and perceptual losses can be used in conjunction with our approach for refining the novel views~\cite{karras2018progressive,wang2018pix2pixHD}.
See supplementary material for further results from other categories.

\subsection{Discussion}
\label{sec:discuss}
Our image to spherical cross-domain embeddings show quantitative improvements over state-of-art in relative 3D object pose estimation. Most existing literature shows results on a restricted set of rotations, and our numbers on 2DOF rotations are comparable to state-of-art. However, for full 3DOF rotations, relative pose estimation from 2D images is especially challenging for approaches which attempt to predict the pose directly from an image embedding, since it requires a combinatorially large training dataset. In contrast, our approach learns the mapping using fewer viewpoints and the corresponding spherical embeddings.

KeypointNet~\cite{suwajanakorn2018discovery} training failed to converge or converged to a bad model for cars 2DOF (noted 'div' in the table) and for the challenging 3DOF rotations.
We found that KeypointNet converges if we limit the 3DOF setting to views on a hemisphere (instead of the full sphere).
Our numbers for the full 3DOF space of rotations are still superior to KeypointNet's results for the limited 3DOF hemisphere.

Evaluation of the \emph{airplane} class is problematic on ShapeNet due to the presence of symmetric instances (flying wings and some spaceships),
which admit two possible alignments that differ by a 180 deg rotation.
We also observe problems on ObjectNet3D, but in this case it's an approximate symmetry that sometimes is not captured by the low resolution spherical CNN feature maps.
In both cases we consider the symmetry when evaluating the errors by making $\text{err}_{\text{sym}} = \min(\text{err}, \pi-\text{err})$.
This metric is used for all methods on \emph{airplanes}.
Note that \citet{suwajanakorn2018discovery} also observe errors around 180 deg and benefit from this metric.
ModelNet40 \emph{airplanes} do not suffer from this issue.

Our method is capable of synthesizing any new viewpoint from any other given viewpoint for any instance of the class it was trained on. 
The categories with less appearance variation are easier to learn and produce sharper images.
For all classes, however, we can verify that the full 3D information is captured by the embeddings.

\section{Conclusion}
\label{sec:conclusion}
In this paper, we explore the problem of learning expressive 3D rotation equivariant embeddings for 2D images. We proposed a novel cross-domain embedding that maps 2D images to spherical feature maps generated by spherical CNNs trained on 3D shape datasets. In this way, our cross-domain embeddings encode images with sufficient shape properties and an equivariant structure that together are directly useful for numerous tasks, including relative pose estimation and novel view synthesis.

We highlight two important areas for future work. First, the cross-domain embedding architecture is composed of a large encoder-decoder structure. The capacity of such a model is greater than what would be necessary for training traditional task-specific models (e.g. a relative pose regression network). This is due to the fact that we are solving a much harder problem: our model must learn a very expressive feature representation that can generalize to many applications. Nonetheless, in future work, we will explore ways to make this component more compact. Second, by construction, our work is tied to the spherical CNNs we use to supervise our embeddings. We will explore alternative rotation equivariant models supervise our training.

Going forward we will also try to improve our embedding representation so that they can be useful for even more challenging tasks such as textured view synthesis for example.

\section{Acknowledgements}
K. Daniilidis is grateful for support through these grants: NSF-IIP-1439681 (I/UCRC), NSF-IIS-1703319, NSF MRI 1626008, ARL RCTA W911NF-10-2-0016, ONR N00014-17-1-2093, ARL DCIST CRA W911NF-17-2-0181, the DARPA-SRC C-BRIC, and by Honda Research Institute.

\bibliography{bib.bib}
\bibliographystyle{icml2019}

\appendix
\begin{figure}[h!]
\centering
\includegraphics[width=\linewidth]{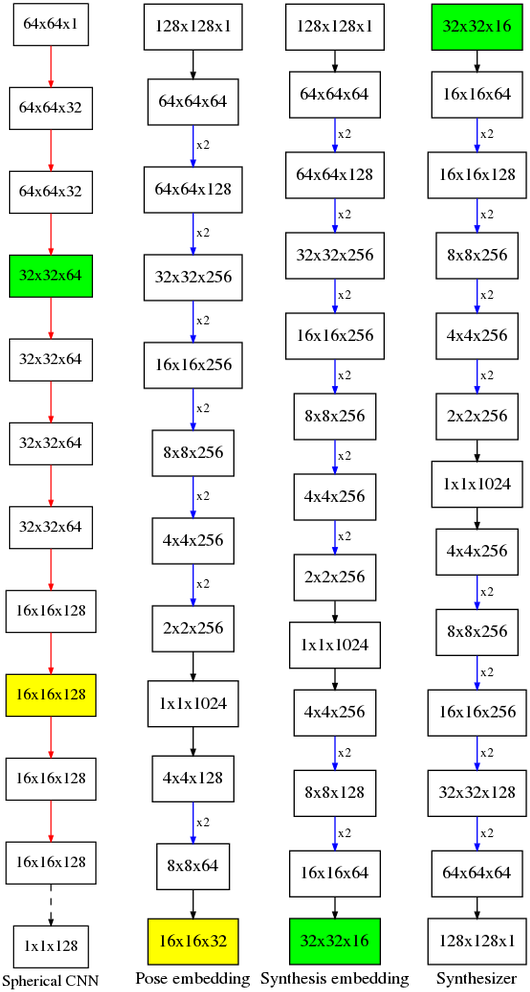}%
\caption{
  \textbf{Network architectures used in this work.}
  Rectangles indicate data dimensions (width x height x channels).
  \textit{Red arrow:} spherical convolutional residual bottleneck layer;
  \textit{dashed arrow:} global average pooling;
  \textit{blue arrow:} residual bottleneck layer;
  \textit{black arrow:} convolutional layer.
  Nodes with yellow and green backgrounds are the target embeddings for pose and synthesis, respectively.
}
\label{fig:layers}
\end{figure}

\section{Architecture details}
\subsection{Spherical CNN}
We train a single 10-layer spherical CNN for object classification on ModelNet40 and use it to obtain the target embeddings for all experiments in this paper.
The basic block is the spherical convolutional residual layer.
The architecture is shown in figure~\ref{fig:layers}, where the a cross-entropy loss over 40 classes is optimized.
The network $s$ is trained for 15 epochs with a batch size of 16, and ADAM optimizer with initial learning rate of $10^{-3}$ which is reduced to $2*10^{-4}$ and $4*10^{-5}$ at steps 5000 and 8500, respectively.
Random anisotropic scaling is used as augmentation.
It achieves 84.2\% accuracy.
\citet{esteves18eccv} achieve 86.9\% on the same task, but with a different architecture containing an extra branch to process surface normals; our inputs are only the ray lengths from the ray casting procedure.

\subsection{Embedding  network}
Our embeddings are obtained with encoder-decoder residual networks.
Given an input with dimensions $N \times N$, the encoding step contains one $7\times 7$ convolutional layer followed by $\log_2 N - 1$ blocks of 2 residual layers, followed by a final convolutional layer that produce a 1D latent vector.
The number of channels double at each residual block, starting at 64 and capped at 256.
Downsampling is through strided convolutions.

The 1D encoding is upsampled using a convolutional layer followed by a sequence of residual blocks and a final $7\times 7$ convolutional layer up to the desired resolution and number of channels, which is $16\times 16 \times 32$ for the pose experiments, and $32\times 32 \times 16$ for novel view synthesis.
Upsampling is through transposed convolutions.

Note that our targets are spherical CNN features inside the residual bottlenecks, so the embeddings have 4 times fewer channels than the actual spherical CNN layer outputs.
The image inputs are $128\times 128$ and 1024 units are used in the 1D encoding.
See figure~\ref{fig:layers} for the resolutions and number of channels per layer.

The embedding network $f$ is trained to minimize a Huber loss for 200k steps with a batch size of 16, and ADAM optimizer with initial learning rate of $2*10^{-4}$ which is reduced to $4*10^{-5}$ and $10^{-5}$ at steps 80k and 180k, respectively.
Random anisotropic scaling of meshes prior rendering is used as augmentation.

\subsection{Synthesis  network}
The synthesizer network $g$ follows the same structure as the embedding, the difference being that the inputs are $32\times 32 \times 16$ and the outputs $128\times 128$.
One question that arises is if the synthesizer should be trained with the target spherical CNN  embeddings as inputs ($s(x)$) or from the embeddings obtained from single views by our network ($f(y)$).
We found that the latter is slightly better.

The synthesis network is trained to minimize an $L_2$ loss for 200k steps with a batch size of 8, and ADAM optimizer with initial learning rate of $2*10^{-4}$ which is reduced to $4*10^{-5}$ and $10^{-5}$ at steps 80k and 180k, respectively.
Random anisotropic scaling of meshes prior to rendering is used as augmentation.

\section{Evaluation details}
In this section, we include details of the training setup for competing approaches.
Note that we still outperform these methods even when allowing more information in the form of oriented meshes, pose supervision and warm starting from pre-trained networks.
\subsection{Regression}
We utilize the same architecture as in the middle columns of figure~\ref{fig:layers} up to the $1024$ dimensional bottleneck, followed by the pose network from \citet{mahendran17cvprw}. We train for $200$k steps, with a batch size of $16$, and ADAM optimizer, with initial learning rate of $1*10^{-4}$, which is reduced to $5*10^{-5}$, $2*10{-5}$, and $8*10^{-6}$ at steps 40k, 75k and 125k respectively.
The RMSE and geodesic loss scheduling similar to \citet{mahendran17cvprw} is used - RMSE loss is used for 100k steps, followed by geodesic loss. When training the 3DOF model, we found that the performance improves when warm starting from a network pre-trained on the 2DOF training set.

The original model is trained to regress a canonical pose.
In our setting, where we render views from a mesh dataset, the meshes need to be aligned or have annotated poses.
Since our method does not require aligned meshes, a more fair comparison would be to train the regression model on pairs of views where the regression target is the relative pose.
We experimented with several variations of this approach and the performance was worse than the regression to a canonical orientation, so we only report results computing the relative pose from the regressed canonical orientations.
\subsection{KeyPointNet}
We utilize the authors' publicly available code and default parameters with minor modifications.
The required changes are because \citet{suwajanakorn2018discovery} distribute the training, which allows a larger batch size of 256, while we train only on a single GPU with a batch size of 24.
We found that with a smaller batch size the default orientation prediction annealing steps (30k-60k) prevents convergence; we changed it to 120k-150k and increased the number of steps from 200k to 300k to be able to reproduce (and slightly improve) the numbers reported in \citet{suwajanakorn2018discovery} (see table~\ref{tab:kptnet}).
We also modify their rendering code to generate the 2DOF and 3DOF datasets, as the original paper only considers a 2DOF hemisphere.

\begin{table}[ht]
  \begin{center}
    \begin{tabular}{|l|rrr|}
      \hline
      & airplane & car & chair\\
      \hline
      Our parameters & 6.06 & 3.31 & 4.94\\
      Original parameters  & 5.72 & 3.37 & 5.42\\
      \hline
    \end{tabular}
  \end{center}
  \caption{Median angular error in degrees for instance based 2DOF hemisphere alignment on ShapeNet.
    Our parameter selection slightly outperforms the original results from \citet{suwajanakorn2018discovery}.}
\label{tab:kptnet}
\end{table}

\section{Extra experiments}
We evaluate image to mesh alignment on ShapeNet and relative pose estimation on ModelNet40.
For completeness, we also include regression results to estimate error to a canonical pose.
Table~\ref{tab:m403d} shows the results for 3DOF alignment and Table~\ref{tab:m402d} for 2DOF alignment for ModelNet40.
\subsection{Image to mesh alignment}
Although we focus on tasks where the inputs are 2D images, our method produces a common equivariant representation for images and meshes that can be used to align images to meshes.
Table~\ref{tab:im-mesh} shows the results.

\begin{table}[ht]
  \centering  
\begin{tabular}{|l|l|c|c|c|c|}
\hline
\multicolumn{2}{|c|}{} & airplane & car & chair & sofa\\
\hline
\multirow{2}{*}{2DOF}  & im-mesh & 5.65 & 4.95 & 13.28 & 12.34\\
 & im-im & 6.24 & 4.73 & 12.10 & 10.80\\
\hline
\multirow{2}{*}{3DOF} & im-mesh & 5.98 & 4.24 & 13.21 & 11.43\\
 & im-im & 7.27 & 4.59 & 12.30 & 9.66\\
\hline
\end{tabular}
\caption{Image to mesh alignment experiment on ShapeNet.
  We show the category based median relative pose error in deg for image to image (\emph{im-im}) and image to mesh (\emph{im-mesh}).}
\label{tab:im-mesh}
\end{table}

\subsection{ModelNet40 relative pose}
Tables \ref{tab:m403d} and \ref{tab:m402d} show alignment results for ModelNet40.
\begin{table}[ht]
  \centering  
  \normalsize
  \setlength\tabcolsep{3pt}
  \begin{tabular}{|l|l|c|c|c|c|c|c|}
    \hline
                      \multicolumn{2}{|c|}{} & airplane & bed           & chair         & car           & sofa          & toilet                        \\ 
    \hline
    \multirow{2}{*}{IB}                      & Regr.    & 11.8          & 26.0          & 43.7          & 16.5          & 25.3          & 17.8          \\
                                             & Ours     & \textbf{7.23} & \textbf{4.93} & \textbf{7.79} & \textbf{3.95} & \textbf{6.51} & \textbf{5.17} \\
    \hline
    \multirow{2}{*}{CB}                      & Regr.    & 12.9          & 29.9          & 52.5          & 15.2          & 34.5          & 17.8          \\
                                             & Ours     & \textbf{8.81} & \textbf{8.55} & \textbf{15.3} & \textbf{5.12} & \textbf{11.0} & \textbf{10.9} \\
    \hline    
    \end{tabular}   
    \caption{Median angular error in degrees for instance (IB) and category-based (CB) 3DOF alignment on ModelNet40.
    }
\label{tab:m403d}
\end{table}

\begin{table}[ht]
\centering
\normalsize
\setlength\tabcolsep{3pt}
\begin{tabular}{|l|l|c|c|c|c|c|c|}
    \hline
      \multicolumn{2}{|c|}{} & airplane   & bed  & chair & car  & sofa & toilet      \\ 
    \hline
    \multirow{2}{*}{IB}      & Regr. & 6.29 & 12.7  & 25.5 & 6.84 & 12.5 & 9.76 \\
                             & Ours       & \textbf{3.33} & \textbf{4.46}  & \textbf{7.07} & \textbf{4.12} & \textbf{4.52} & \textbf{4.88} \\
    \hline
    \multirow{2}{*}{CB}      & Regr. & 7.13 & 15.8  & 32.2 & 7.00 & 13.3 & \textbf{10.4} \\
                             & Ours       & \textbf{4.80} & \textbf{6.60}  & \textbf{10.2} & \textbf{4.82} & \textbf{9.56} & 10.8 \\
    \hline
    \end{tabular}   
  \caption{Median angular error in degrees for instance (IB) and category-based (CB) 2DOF alignment on ModelNet40.}
\label{tab:m402d}
\end{table}

\subsection{Novel view synthesis}
\begin{figure*}
\centering
\includegraphics[width=\linewidth]{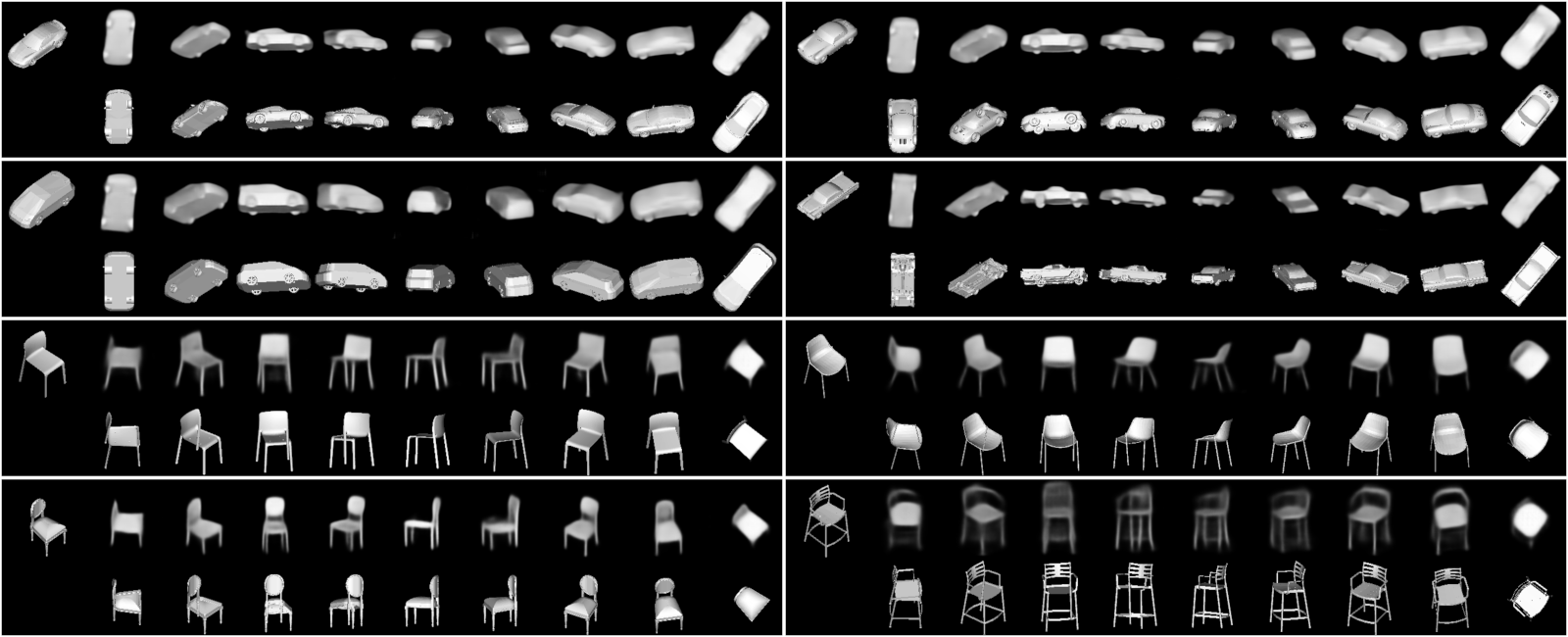}%
\caption{
  \textbf{More novel view synthesis results.}
  \textit{Top-left:} inputs, which are 2D images from the test set. 
  \textit{Top row:} novel views generated using our method.
  \textit{Bottom row:} ground truth views rendered from the original mesh.
  The bottom right shows a failure case due to a chair with uncommon appearance.
}
\label{fig:supp_synth}
\end{figure*}

We show novel view synthesis results for other classes, including a failure case in~\ref{fig:supp_synth}.

\subsection{Visualization}
\begin{figure*}
\centering
\includegraphics[width=\linewidth]{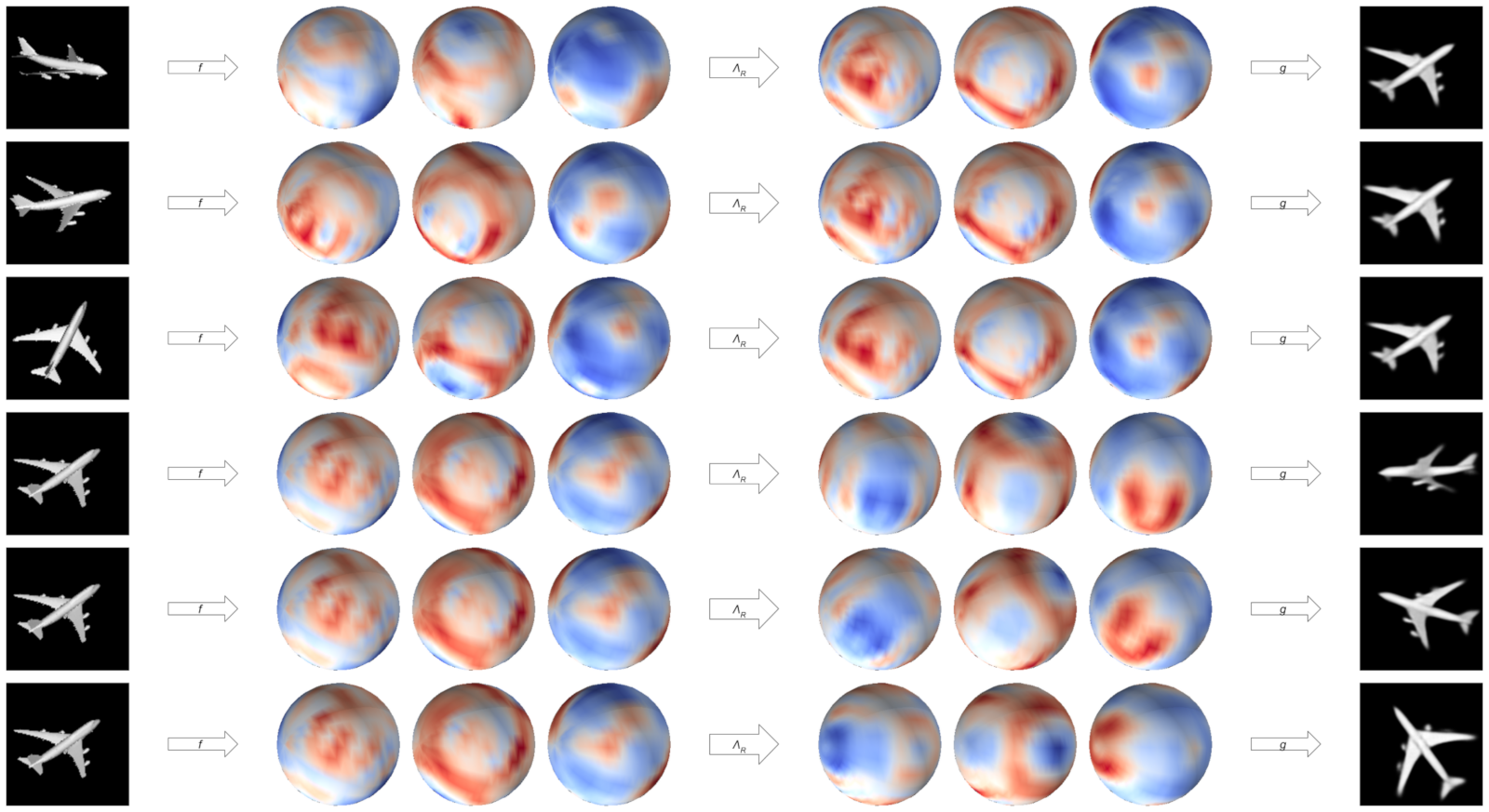}%
\caption{
  \textbf{Novel view synthesis visualization.}
  \textit{Each row:} inputs, 3 embedding channels, rotated embedding channels, outputs.
  Top 3 rows show generation of a canonical view from arbitrary views and correspond to \texttt{synthesis1.gif}.
  Bottom 3 rows show generation of arbitrary views from a canonical view and correspond to \texttt{synthesis2.gif}.
}
\label{fig:synth_anim_frames}
\end{figure*}

\begin{figure*}
\centering
\includegraphics[width=\linewidth]{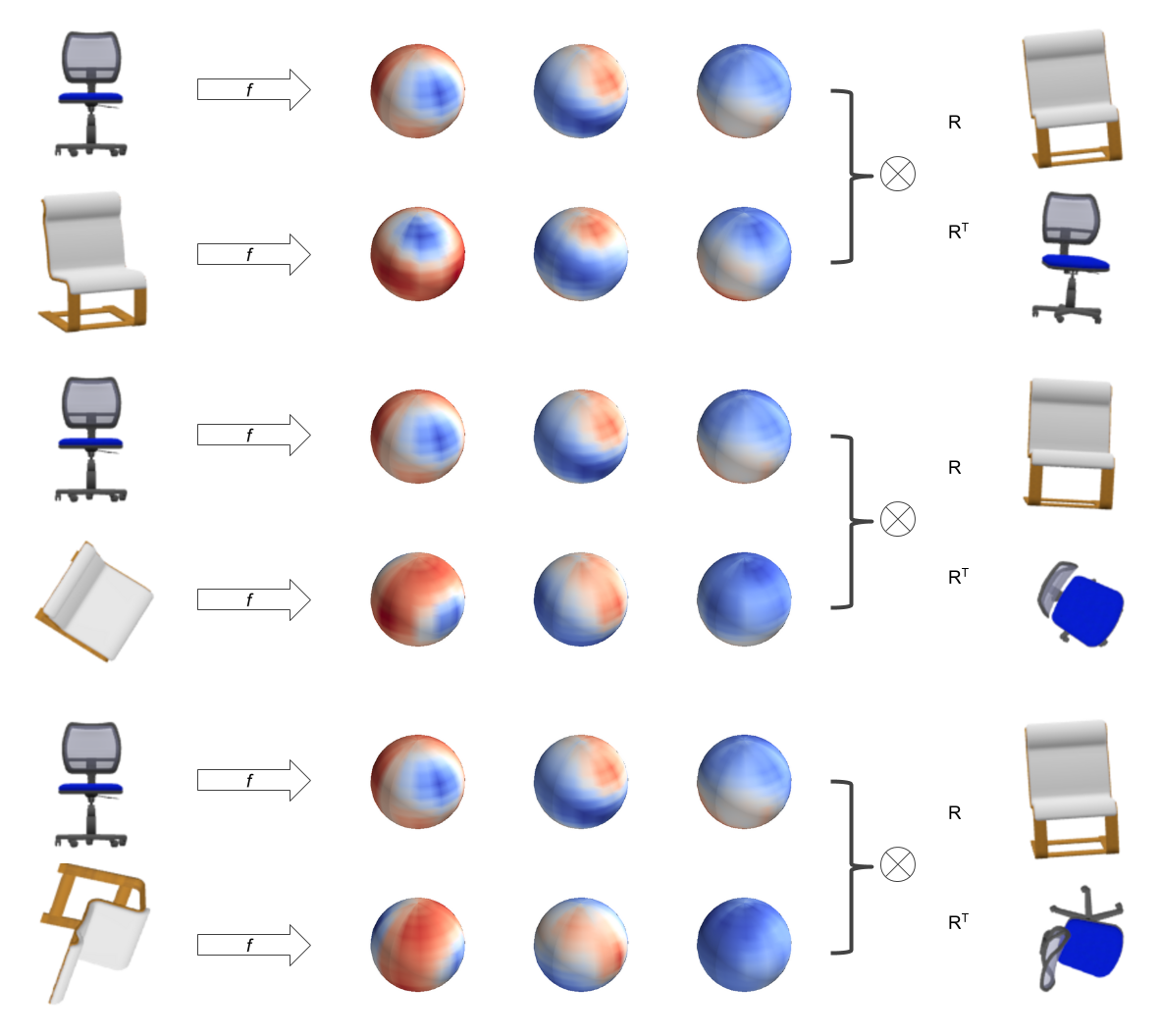}%
\caption{
  \textbf{Relative pose estimation visualization.}
  \textit{Each block of two rows:} pair of inputs, 3 embedding channels per input, mesh 2 rotated into pose 1, and mesh 1 rotated into pose 2.
  We render from the ground truth meshes for visualization purposes only; our inputs are solely the 2D views and output is the relative pose.
  See \texttt{pose.gif} for an animation.
}
\label{fig:pose_anim_frames}
\end{figure*}

Figure~\ref{fig:synth_anim_frames} shows inputs, embedding channels, rotated embedding channels and outputs from synthesis.
Figure~\ref{fig:pose_anim_frames} shows inputs, embedding channels, and alignment visualization.
See animations in the supplementary material (\verb|synthesis1.gif|, \verb|synthesis2.gif|, \verb|pose.gif|).

\end{document}